\documentclass[letterpaper, 10 pt, conference]{ieeeconf}  

\IEEEoverridecommandlockouts                              
\usepackage{booktabs}
\usepackage{graphicx}
\usepackage{flushend}
\usepackage{mathtools}
\usepackage{amsmath}
\usepackage{environ}
\NewEnviron{myequation}{%
    \begin{equation}
    \scalebox{1.1}{$\BODY$}
    \end{equation}
    }

\title{\LARGE \bf
Digital Image Forensics using Deep Learning
}

\author{Akash Nagaraj*, Mukund Sood, Vivek Kapoor, Yash Mathur, Bishesh Sinha\\
Department of Computer Science, \\
PES University, Bangalore, India \\
{\tt\small [akashn1897, mukundsood2013, vivekkapoor2959, mathuryash5, bisheshsin]@gmail.com}
}

\begin{document}

\maketitle
\thispagestyle{empty}
\pagestyle{empty}

\begin{abstract}

During the investigation of criminal activity when evidence is available, the issue at hand is determining the credibility of the video and ascertaining that the video is real. Today, one way to authenticate the footage is to identify the camera that was used to capture the image or video in question. While a very common way to do this is by using image meta-data, this data can easily be falsified by changing the video content or even splicing together content from two different cameras. Given the multitude of solutions proposed to this problem, it is yet to be sufficiently solved. The aim of our project is to build an algorithm that identifies which camera was used to capture an image using traces of information left intrinsically in the image, using filters, followed by a deep neural network on these filters. Solving this problem would have a big impact on the verification of evidence used in criminal and civil trials and even news reporting.

\end{abstract}
\vspace{0.25 cm}
\textbf{\textit{ Keywords- source camera; fourier transform; deep learning}}


\section{Introduction}
Images have become a significant source of information in the modern era. The advent of the Internet and the increasing volume of data and storage facilities has further aided the ease of circulation of images to a large section of the population, be it friends, family or acquaintances. This problem leads to us often finding ourselves in situations where we question the credibility of a certain image or video. At face value, this may seem inconsequential in most situations, however, this gives power to a large number of people to spread false news and doctored images to suit their own objectives. \par
In this day and age, the advent of many social networking sites has made it imperative to validate images and their sources, to ensure they are authentic and not a false representation of the facts. \par
Most of the new approaches to solve this problem have been very computationally expensive and have required large amounts of data. They use models on large amounts of data with no preprocessing which results in a very brute-force approach to solving this problem. These models with a enough data will be able to pick up on trace amounts of information enabling accurate classification. \par
However, we have taken a contrasting approach. We worked with a very small amount of data (2750 images in total) and took a more technical approach to solving the problem. We worked on explicitly extracting information embedded in an image instead of relying purely on the ability of the network to do so. We explore different forms of information in this paper such as finding Source Pattern Noise (SPN) information, probability plots of noise, detection of image interpolation and even extracting information using a Gray Level Dependency Matrix~\cite{c1}. During our experiments, some of these methods worked exceedingly well while others led to dead ends, as further described. \par
We are currently working on creating an ensemble of these models to give us the best result. The work we have done so far has given us fairly promising results however we believe that with some more fine-tuning it will be a very competitive model. \par
This objective of this project was to solve the multi-faceted problem described in the abstract. We aim to prevent falsification of facts and the propagation of fake images or videos through the Internet. \par


\section{Related Work}

\subsection{A Survey on Digital Camera Image Forensics Methods~\cite{c2}}
Using sensor imperfections such as Sensor Pattern Noise detection provides us with a reliable method for identifying the source camera. The photo-response-non-uniformity (PRNU) where different pixels have different light sensitivities due to imperfections in the sensor manufacturing process is a major source of pattern noise. This makes PRNU a natural feature for uniquely identifying sensors.\par
Exploring the Camera Filter Array (CFA) Interpolation process to determine the correlation structure present in each colour band, can also be used for image classification. The main assumption is that the interpolation algorithm and the design of the CFA filter pattern of each manufacturer (or even each camera model) are somewhat different from others, which will result in a distinguishable correlation structure in the captured images.

\subsection{Deep Learning for Source Camera Identification on Mobile Devices~\cite{c3}}
The authors of this paper designed a Convolutional Neural Network Architecture using three Convolutional Layers with a Rectified Linear Units activation function followed by a Max Pooling Layer, followed by the Classification which comprised of three Fully Connected Layers followed by a Softmax Layer. The primary principle  behind the method used in this paper, is that there was no preprocessing used, raw image data was passed to the CNN, and the image is first passed through a High Pass Filter (5x5 Kernel) and the filtered image is cropped to a standard size of 252x252 pixels, and then fed to the CNN.

\subsection{Experiments on Improving Sensor Pattern Noise Extraction for Source Camera Identification~\cite{c4}}
The authors of this paper believe that the effectiveness of the identification technique when using SPN Extraction depends largely upon the quality of SPN being used. The SPN extraction procedure may perform bad when applied, for example, to images where the details of a scene are not eliminated by the de-noising filter, because of their intensity. They provide a solution to this issue by providing methods to enhance SPN extraction such as using only large components of Sensor Pattern Noise or enhancing SPN Extraction by trying to attenuate the strongest components of the SPN, since it is observed that the stronger the signal component is in a sensor pattern noise, the more likely it is to be associated to strong scene details and hence, the less trustworthy that component should be. 

\subsection{Source Camera Identification using GLCM~\cite{c5}}
GLCM (Gray Level Dependency Matrix) is used to extract features such as Entropy, Contrast, Homogeneity, Correlation etc. from an image. The authors of this paper first implemented Image Sharpening using Hybrid Edge Detection, then estimating the SPN followed by extracting GLCM features from the SPN. This information was then stored in a database, and their model of prediction was finding the correlation between test images, and the features stored in the database. 


\section{Approach}

\subsection{The Dataset}
The dataset we are using for this project is The IEEE Signal Processing Society - Source Camera Model Identification.~\cite{c0} \par
About the data set:
\begin{itemize}

\item The dataset consisted of 10 different camera models, namely:
\begin{itemize}
\item Samsung Galaxy S4
\item Samsung Galaxy Note 3
\item Sony NEX-7
\item LG Nexus 5x
\item HTC One M7
\item Motorola Moto X
\item Motorola Nexus 6
\item Motorola DROID MAXX
\item Apple iPhone 6
\item Apple iPhone 4s

\end{itemize}
\item The training set consists of 275 images taken by each camera model to be classified; resulting in a total of 2750 images. 
\item The test set consists of 2000 images taken by the same models with half of the images in the test set being altered. The set of possible processing operations that were performed are: 
    \begin{itemize}
    \item JPEG compression with quality factor = 70 or 90
    \item Resizing (via bicubic interpolation) by a factor of 0.5, 0.8, 1.5 or 2.0
    \item Gamma correction using gamma = 0.8 or 1.2
    \end{itemize}
\item None of the images in the test data were taken with the same device as in the train data.

\end{itemize}

\subsection{Evaluation Metric}

\begin{center}
\begin{myequation}
weightedAccuracy(y_{i}, \hat{y_{i}})  = \frac{1}{n} \sum_{i=1}^{n} \frac{w_{i}(y_{i} \hspace{1pt}=\hspace{1pt} \hat{y_{i}})}{\sum_{} w_{i}}
\end{myequation}
\end{center}
The evaluation is based on the weighted categorization accuracy of our predictions (the percentage of camera models correctly predicted). In the above equation, \textit{n} is the number of samples in the test set, \textit{y} is the true camera label, \textit{y\underline{ }hat} is the predicted camera label, and \textit{w(i)} is 0.7 for unaltered images, and 0.3 for altered images.


\subsection{Preprocessing}

We built four models; one using an SVM, one using a Convolutional Neural Network, and one which was a combination of the SVM and CNN and the final model was based on the existing ResNet50 Architecture. We had one level of common preprocessing for all four models, as described below. \par
After the common layer of preprocessing, the SVM, the CNN and the combination of the two had a common second layer of preprocessing, however, we used a unique preprocessing technique for the ResNet model. \par
In general, for all the images, we preprocessed them by manipulating using:
\begin{itemize}
\item JPEG compression with quality factor = 70
\item JPEG compression with quality factor = 90
\item gamma correction using gamma = 0.8
\item gamma correction using gamma = 1.2
\item resizing (via bicubic interpolation) by a factor of 0.5
\item resizing (via bicubic interpolation) by a factor of 0.8
\item resizing (via bicubic interpolation) by a factor of 1.5
\item resizing (via bicubic interpolation) by a factor of 2.0
\end{itemize}
These manipulations were performed to random images, and the training data consisted of both untouched, and manipulated images to ensure the model we built is a robust one. 

Our preprocessing consisted of extracting three sources of information from each image: Source Pattern Noise, Image Interpolation and GLCM.\par 
To extract Source Pattern Noise, we passed the original image through a high-pass filter (Bilateral Filter) which resulted in a de-noised image. We then subtracted the de-noised image from the original image to obtain the noise for the image. We took only the top 4\% of values because we theorized that lower values of noise would contain common noise information that could be obtained from images across all camera models. The top 4\% would give us a better estimate of the noise values for the particular camera model that we were dealing with. This procedure was followed for every image of the dataset.\par
\begin{figure}[h]
\caption{Noise Pattern plots in the Samsung Galaxy S4}
\centering
\includegraphics[width=\linewidth]{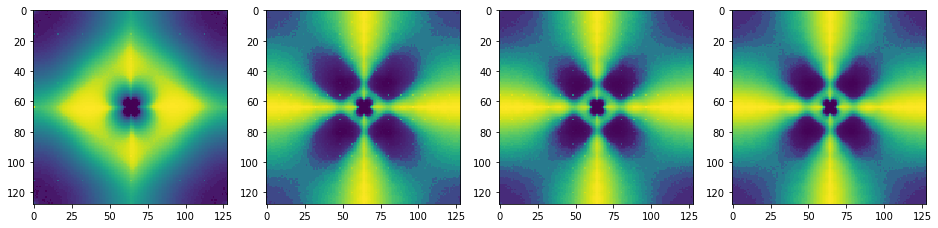}
\caption{Noise Pattern plots in the Sony NEX-7}
\centering
\includegraphics[width=\linewidth]{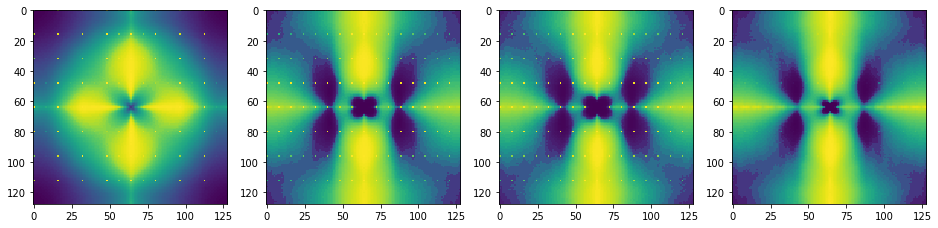}
\end{figure}
To detect image interpolation, we followed the Gallagher and Chen Algorithm. In their algorithm, Gallagher and Chen look at only the green channel, however,  they state that other channels could also be considered if taken individually. First, they convolve the image with a high pass filter to remove low-frequency information (which is irrelevant) and then calculate variance across diagonals to form a single dimensional signal of variances. (Diagonals are used because it is clear from the pattern that the variance across diagonals vary periodically). For an interpolated image, we expect to find periodicity in this signal; Gallagher and Chen used a Discrete Fourier Transform. For CFA Interpolation, we expect to see a peak in the DFT. Like the previous method, we performed this procedure for every image in the dataset.\par
To extract GLCM Features, we passed the original image through a Canny Filter and simultaneously through a Laplacian Filter separately. We compared each pixel value of the two resulting images and formed a new image that comprised of the maximum pixel values. This new image was subtracted from the original image. We applied thresholding to the image, and converted it to gray-scale. This final image was then used to extract GLCM features using the Gray Level Dependency Matrix.\par

\subsection{Preprocessing for ResNet50 [6]}

We describe a novel approach to this problem here. Our initial approach was to simply pass the image through a high pass filter, and pass the resulting image to ResNet to see if it would be able to find intrinsic patterns in the image. However, this did not give us very high accuracies in terms of accuracy. While the results were promising, our next step was to make a Probability Plot of the noise in each image, and save it as a .jpg image. Our method was simple; given a single image, we took a random 256x256 crop of the image, and applied a Gaussian Blur Filter to the image to remove noise. Once noise was removed, we subtracted the original crop from the Filtered crop to obtain the noise in that crop. Next, we applied a Fast Fourier Transform to the resulting crop, with four different shifts, to obtain four distinct images for each image. We did this 256 times for each shift for each image, and finally the result of the 256 random crops was averaged out to give us the four images. \par


\section{Experimental Results}

\subsection{Support Vector Machine model}
The SVM model used consisted of two Layers. In the first layer, we built 10 binary SVMs, one for each camera model. Each SVM would be used to classify an image as either \textit{Camera X} or \textit{Not Camera X}. Now, if only one SVM out of the 10, classified an image as \textit{Camera X}, then this image would receive a final classification of \textit{Camera X}. However, if two or more SVMs classified that image as being part of its class, then we would train a Multi-Class SVM on the training data pertaining to the SVMs that tested positive. In the second case the final classification obtained from the multi-class SVM would be considered as the final classification of the image. \par

\subsection{Convolutional Neural Network model}
The CNN model used was similar to the one used in camera model identification With The Use of Deep Convolutional Neural Networks. The architecture for the network used was similar. The preprocessing used however, was entirely different (as mentioned in the previous section). The CNN was made of three Convolutional Layers, followed by a Max Pooling Layer and three Fully Connected Layers. The final Activation function used was a Softmax function. Each image was individually preprocessed using the above mentioned method, and used to train the CNN model. \par

\subsection{ResNet-50~\cite{c6}} 
We used the images subject to the preprocessing described above to train the ResNet50 model. In general, in a deep convolutional neural network, multiple layers are stacked together and are trained to the perform task at hand. The network learns several low/mid/high level features at the end of its layers. In the case of residual learning, instead of trying to learn some features, we try to learn some residual. Residual can be simply understood as subtraction of feature learned from input of that layer. ResNet does this using shortcut connections (directly connecting input of nth layer to some $(n+x)^{th}$ layer. It has proved that training this form of networks is easier than training simple deep convolutional neural networks and also the problem of degrading accuracy is resolved. \par
We have used ResNet50 to do the actual predictions of our models based on the noise patterns of the camera's sensor, by preprocessing the images.

\subsection{Results}
We tested our model on the test set provided by the IEEE Signal Processing Society sponsored competition on Kaggle. Our current accuracy stands at 87\%, which is promising, especially when compared to the other methods used for this problem which are a lot more computationally intensive, and require large amounts of data. Another unique feature of the model we have proposed is that it has can successfully classify between 10 different classes, and can further be extended to a larger number of classes.\par


\section{Conclusions}
Our work so far has led to promising results, however we believe that by extracting some more information from each image and streamlining the information used, it can lead to better and more robust results. We believe that since we explicitly extract relevant information about each image, our model is a more robust one as opposed to the other common brute force approaches. \par

\section{Reproducible Research}
In the spirit of reproducible research, the work done is publicly available. The Python code for all the experiments and work done in this paper can be accessed at~\cite{c7}.

\addtolength{\textheight}{-12cm}   


\end{document}